\documentclass[conference,letterpaper,twoside]{IEEEtran}
\usepackage[inner=0.625in, outer=0.625in, top=0.75in, bottom=1in]{geometry}
\IEEEoverridecommandlockouts
\usepackage{cite}
\usepackage{tabularx}
\usepackage{adjustbox}
 \usepackage{graphicx}
\usepackage{multirow}
\usepackage{textcomp}
\usepackage{subcaption}
\usepackage{colortbl}
\usepackage{hhline}
\usepackage{rotating}
\usepackage[utf8]{inputenc}
\usepackage{algorithm}
\usepackage{algpseudocode}
\usepackage[draft,hidelinks,colorlinks=true,citecolor=blue]{hyperref}
\usepackage[table,xcdraw,dvipsnames]{xcolor}
\usepackage{tabulary}
\usepackage{amsmath,amssymb,amsfonts,amsthm}
\usepackage{array}

\allowdisplaybreaks
\usepackage{tikz}
\usepackage{tikz-cd}
\usepackage{stackengine}
\usepackage{scalerel}
\usepackage{fp}
\def\clipeq{\!\mathrm{=}\!}
\def\Lla{\Longleftarrow}
\def\Lra{\Longrightarrow}
\newcount\argwidth
\newcount\clipeqwidth
\savestack\tempstack{\stackon{$\clipeq$}{}}%
\clipeqwidth=\wd\tempstackcontent\relax
\newcommand\xleftrightarrows[2]{%
\savestack\tempstack{\stackon{$\scriptstyle#1$}{$\scriptstyle#1$}}%
\argwidth=\wd\tempstackcontent\relax%
\FPdiv\scalefactor{\the\argwidth}{\the\clipeqwidth}%
  \FPsub\scalefactor{\scalefactor}{.4}
  \FPmax\scalefactor{\scalefactor}{.04}%
  \mathrel{%
  \stackunder[1pt]{\stackon[3pt]{$\Lla\hstretch{\scalefactor}{\clipeq}\Lra$}%
     {$\scriptstyle#1$}}{$\scriptstyle#2$}%
  }%
}
\usepackage{float}
\usepackage{placeins}
\usepackage{enumitem}
\usepackage{changepage}
\theoremstyle{definition}

\def\BibTeX{{\rm B\kern-.05em{\sc i\kern-.025em b}\kern-.08em
    T\kern-.1667em\lower.7ex\hbox{E}\kern-.125emX}}

\begin{document}
\title{Accurate AI-Driven Emergency Vehicle Location Tracking in Healthcare ITS Digital Twin}
\author{
\IEEEauthorblockN{
Sarah Al-Shareeda\IEEEauthorrefmark{1}\IEEEauthorrefmark{2}\IEEEauthorrefmark{4}, Yasar Celik\IEEEauthorrefmark{1}, Bilge Bilgili\IEEEauthorrefmark{1}, Ahmed Al-Dubai\IEEEauthorrefmark{3}, and Berk Canberk\IEEEauthorrefmark{3}}
\IEEEauthorblockA{\IEEEauthorrefmark{1}Department of AI and Data Engineering, Istanbul Technical University, Turkey}
\IEEEauthorblockA{\IEEEauthorrefmark{2}BTS Group, Turkey}
\IEEEauthorblockA{\IEEEauthorrefmark{4}Center for Automotive Research (CAR), The Ohio State University, USA}
\IEEEauthorblockA{\IEEEauthorrefmark{3}School of Computing, Engineering and The Built Environment, Edinburgh Napier University, UK}

Email: \{alshareeda, celiky20, bilgili21\}@itu.edu.tr and \{a.al-dubai, b.canberk\}@napier.ac.uk
}
\maketitle

\begin{abstract}
\textcolor{black}{Creating a Digital Twin (DT) for Healthcare Intelligent Transportation Systems (HITS) is a hot research trend focusing on enhancing HITS management, particularly in emergencies where ambulance vehicles must arrive at the crash scene on time and track their real-time location is crucial to the medical authorities. Despite the claim of real-time representation, a temporal misalignment persists between the physical and virtual domains, leading to discrepancies in the ambulance's location representation. This study proposes integrating AI predictive models, specifically Support Vector Regression (SVR) and Deep Neural Networks (DNN), within a constructed mock DT data pipeline framework to anticipate the medical vehicle's next location in the virtual world. These models align virtual representations with their physical counterparts, i.e., metaphorically offsetting the synchronization delay between the two worlds. Trained meticulously on a historical geospatial dataset, SVR and DNN exhibit exceptional prediction accuracy in MATLAB and Python environments. Through various testing scenarios, we visually demonstrate the efficacy of our methodology, showcasing SVR and DNN's key role in significantly reducing the witnessed gap within the HITS's DT. This transformative approach enhances real-time synchronization in emergency HITS by approximately 88\% to 93\%.}
\end{abstract}

\begin{IEEEkeywords}
Healthcare ITS, Digital Twins, Location Prediction, Artificial Intelligence, Delay Offsetting
\end{IEEEkeywords}
\section{Problem in Focus}\label{intro}
\begin{figure*}[!htbp]
\centering
\includegraphics[width=.6\linewidth]{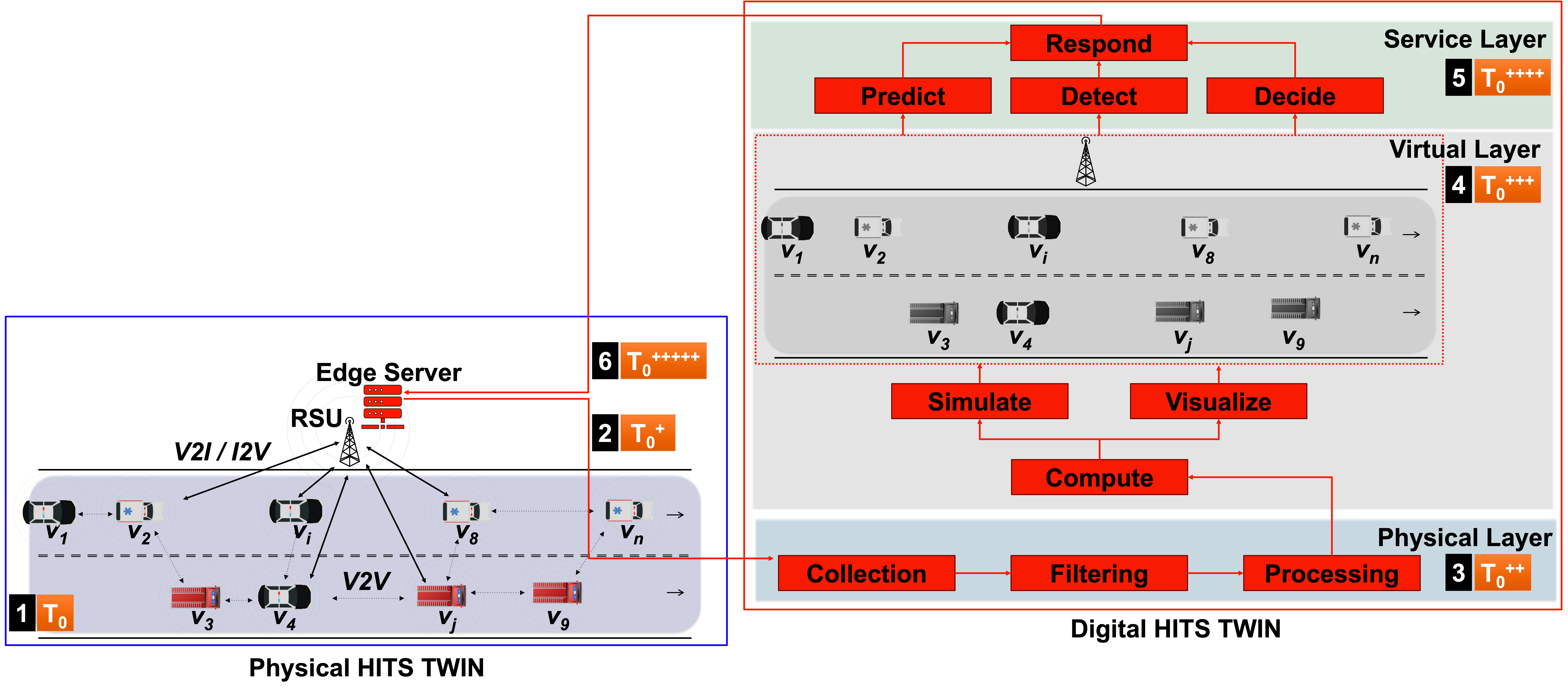}
\caption{Problem in Focus: Witnessed Synchronization Delay between the Physical and Digital Worlds}\label{fig:plan2}
\end{figure*}

In contemporary transportation management, Intelligent Healthcare Transportation Systems (HITS) are fundamental to facilitating collaborative information exchange among medical vehicles and infrastructure, especially in the case of emergencies and accidents, and incorporating Digital Twins (DTs) into HITS promises significant advantages, including real-time data integration, improved traffic management, and democratizing data-driven decision making. DT deployment within HITS can take various forms, including cloud-based, edge-based, or hybrid-based approaches \cite{luan2021paradigm,al2024does,al2025group,saim2024safety,saim2024control}. However, achieving real-time synchronization between the physical and virtual facets of HITS remains elusive, resulting in an enduring temporal gap that hinders the attainment of a real-time representation of the physical system. For example, in edge-based deployment, this delay originates from the process by which the physical HITS transmits information to the edge server that hosts the constructed DT.
Moreover, this temporal discrepancy is observed across the DT's data, virtual, and service layers. Ultimately, when final decisions and responses are transmitted to the physical system, the system's state may have changed before receiving these responses, underscoring the challenge of achieving true real-time synchronization. This temporal misalignment is exemplified in Fig. \ref{fig:plan2}, where the physical HITS records data at time (say $T_0$\footnote{Current timestamp $T_0$, three-times delayed timestamp $T_0^{+++}$, and five-times delayed timestamp $T_0^{+++++}$.}), while the DT lags significantly at delayed timestamp $T_{0}^{+++}$. Consequently, the response is returned at a very late timestamp $T_{0}^{+++++}$. This discrepancy between real-time data acquisition and DT response generation requires a dire solution. This research problem underscores the critical need to bridge the synchronization gap between physical HITS and their DTs, thus advancing the quest for real-time representation of healthcare transportation systems. Numerous studies have addressed the issue of synchronization from various angles \cite{yang2022data, zheng2023data}. An emerging notable approach involves using AI-based predictive models within the DT plane, allowing predicting medical vehicle locations and behaviors ahead of time, effectively mitigating synchronization delays. A detailed literature review reveals a multitude of endeavors addressing the general challenge of location prediction \cite{chen2021origin,dasanayaka2022analysis,miri2022novel,zheng2023resource}. These works collectively showcase the use of AI for vehicle location prediction, each contributing innovative techniques and models to improve prediction accuracy, efficiency, and adaptability. However, few of these works utilize the concept of DT, that is, making location prediction in the DT domain; Maheswaran et al. \cite{maheswaran2019fog} improve autonomous driving systems by continuously updating the locations of autonomous and human-piloted vehicles on road segments. To mitigate communication delays, they incorporate a forecaster algorithm in the twin capable of predicting future vehicle locations. Along the same lines, our present study aims to bridge the gap by contributing to the field of HITS in the following key ways:
\subsection{Main Contributions}
\begin{enumerate}
\item We address the challenge of the temporal gap in HITS DT using AI prediction techniques, ensuring a seamless alignment between virtual and actual medical vehicle locations.
\item We introduce Support Vector Regression (SVR) as our Machine Learning (ML) model and Deep Neural Networks (DNN) as the Deep Learning (DL) model. This approach takes advantage of the strengths of both the ML and the DL techniques to predict the next location of the ambulance accurately.
\item A thorough performance analysis of the SVR and DNN models is conducted using three key metrics in two simulation environments, MATLAB and Python, proving the consistency of the models across different platforms.
\item We build a mock DT environment as a Proof of Concept (PoC) using Docker \cite{Docker} and Apache Kafka \cite{kafka} to support a real-time HITS data pipeline of actual and predicted locations. Our DT enables accurate data visualization via Grafana \cite{grafana}, enhancing synchronization in HITS operations.
\item We showcase the efficacy of our models by demonstrating a significant reduction in the witnessed delay within the HITS DT. This improvement contributes to an improved accuracy of real-time synchronization.
\end{enumerate}

Our research advances HITS by reducing DT witnessed delay, improving prediction accuracy, and offering practical solutions for real-time applications in healthcare transportation systems. The rest of the paper is organized as follows: the preliminary description of our actual and virtual HITS and the proposed prediction models are presented in Section \ref{design}. In Section \ref{results}, the performance of the proposed models is evaluated, the DT is constructed, and the effect of prediction on compensating for the observed delay is discussed. Finally, the paper is concluded with key extensions in Section \ref{conc}.

\section{Proposed Solution: Predicting the next location of the ambulance in HITS DT to offset the observed delay}\label{design}
\begin{figure}[!htbp]
    \centering
    \includegraphics[width=\linewidth]{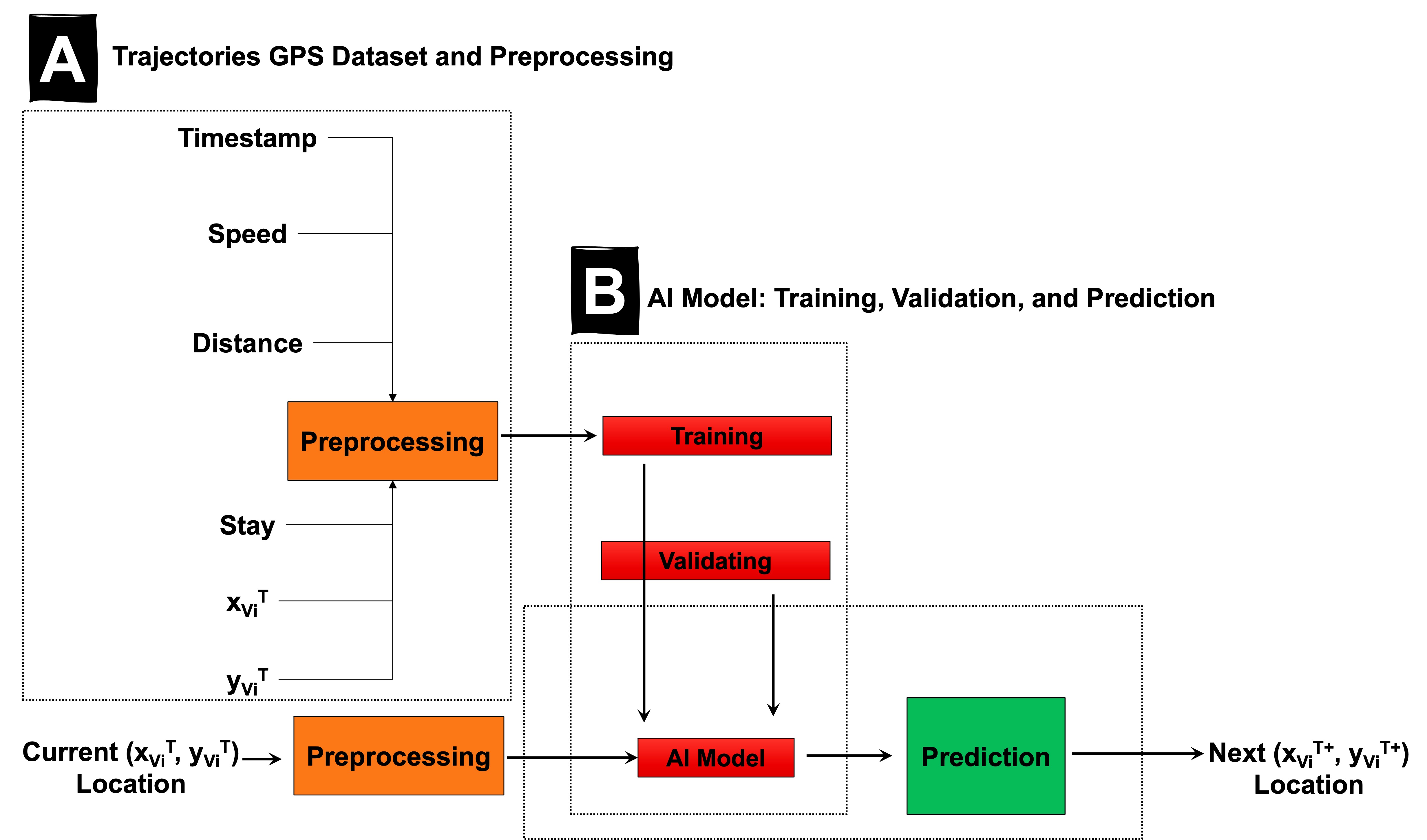}
    \caption{Predictive Model Design Structured Steps Within The DT Layers}\label{fig:plan3}
\end{figure}

Our HITS comprises a fleet of $n$ emergency service vehicles, as depicted in Fig. \ref{fig:plan2}, each identified by the index $V_i$, $i=\{1, ... n\}$. These vehicles are equipped with real-time location tracking systems that provide coordinates $(x_{v_i}^T, y_{v_i}^T)$ at time $T$. Communication of status occurs through vehicle-to-vehicle (V2V) and vehicle-to-structure (V2I) channels, with real-time situational information transmitted to the Road Side Unit (RSU). Despite adjusting for factors such as congestion and contention in the physical network, an unavoidable communication delay $\Delta T$ is observed from the perspective of the RSU. Deploying a virtual twin of the HITS at the RSU's edge server exacerbates the temporal gap and asynchronization with the physical system. The DT of each vehicle consistently lags by $\Delta \tau$ seconds behind its real-world counterpart. Our primary goal is to mitigate this observed $\Delta \tau$. To address this, we integrate AI models to predict future vehicle locations within the DT framework, aligning the virtual representation with physical reality. Two AI models, SVR as an ML model and DNN as a DL model, operate within the DT layers. The schematic representation of the prediction model is displayed in Fig. \ref{fig:plan3} with its three involved steps detailed below.

\subsection{Input Dataset and Preprocessing}
Our dataset comprises a historical geospatial GPS trajectory dataset sourced from a fleet of $n=221$ regular vehicles, which operated during the first two days of each month throughout 2019. This GPS dataset includes the following feature set: unique vehicle identifiers $V_i$, $i=\{1, ..., n\}$, timestamp, speed (km / h), distance traveled (m), duration of stay at a specific location (second), latitude and longitude coordinates of the current location, that is, $(x_{V_i}^T, y_{V_i}^T)$ and the corresponding next location coordinates $(x_{V_i}^{T+}, y_{V_i}^{T+})$. In particular, this data set comprises 1,048,576 timestamps per month; however, to ensure data quality and minimize computational resource utilization, we apply rigorous data filtering and restrict our study to a subset of $N=43,856$ time instances. Subsequently, this filtered data set is divided into a training subset, which constitutes 80\% of the data, and a validation subset, reserving the remaining 20\%. Data preprocessing forms a foundational corner of our model development, adhering to rigorous technical standards. In this critical process, a meticulous transformation of our features set centers the data around a mean of $0$ and scales it to show a standard deviation of $1$. This standardization significantly improves the convergence rate during model training, providing AI models with a notable understanding of the intricate spatio-temporal dynamics of our dataset. This centering will also affect the bias of the SVR model $b$, as explained in a later note. The next stage involves feeding the data into our selected prediction models, enabling us to anticipate vehicle movements accurately. In this context, we choose the SVR and DNN models, as SVR makes precise predictions. At the same time, DNNs can discern intricate spatio-temporal patterns, resulting in highly accurate and dependable forecasts of future vehicle locations. We dive into a detailed exploration of these two models in the following.
\subsection{Utilized AI Prediction Models Description}
\subsubsection{SVR ML Predictive Model}
In the context of our dataset, the main objective of regression is to uncover how the changes in the six input features relate to the changes in the next location coordinates $(x_{V_i}^{T+}, y_{V_i}^{T+})$ of the vehicle. SVR is our dataset's regression method of choice as it can effectively model the sought nonlinear relationship. SVR is an extension of the Support Vector Machine (SVM) algorithm that can capture complex relationships and patterns within the data by finding the ``hyperplane" that best fits the data points within error $\varepsilon$-tube region. In our case, this hyperplane is a mathematical representation of the relationships between the input features and the target x-coordinate $x_{V_i}^{T+}$ and y-coordinate $y_{V_i}^{T+}$. We employ dual SVR models: one dedicated to predicting the x-coordinate and the other to predict the y-coordinate. For simplicity, our description refers to either coordinates as $\hat y$ or $f(\hat x)$ where $\hat x=[\hat x_1, \hat x_2, ..., \hat x_{\hat i}, ..., \hat x_N]$ represents the input dataset that has $N=43,856$ instances of 6-dimension $\hat x_{\hat i}$ values. The main steps of formulating our SVR, exhibited in Fig. \ref{fig:plan4}, involve:
\begin{figure*}[!htbp]
    \centering
    \includegraphics[width=\linewidth]{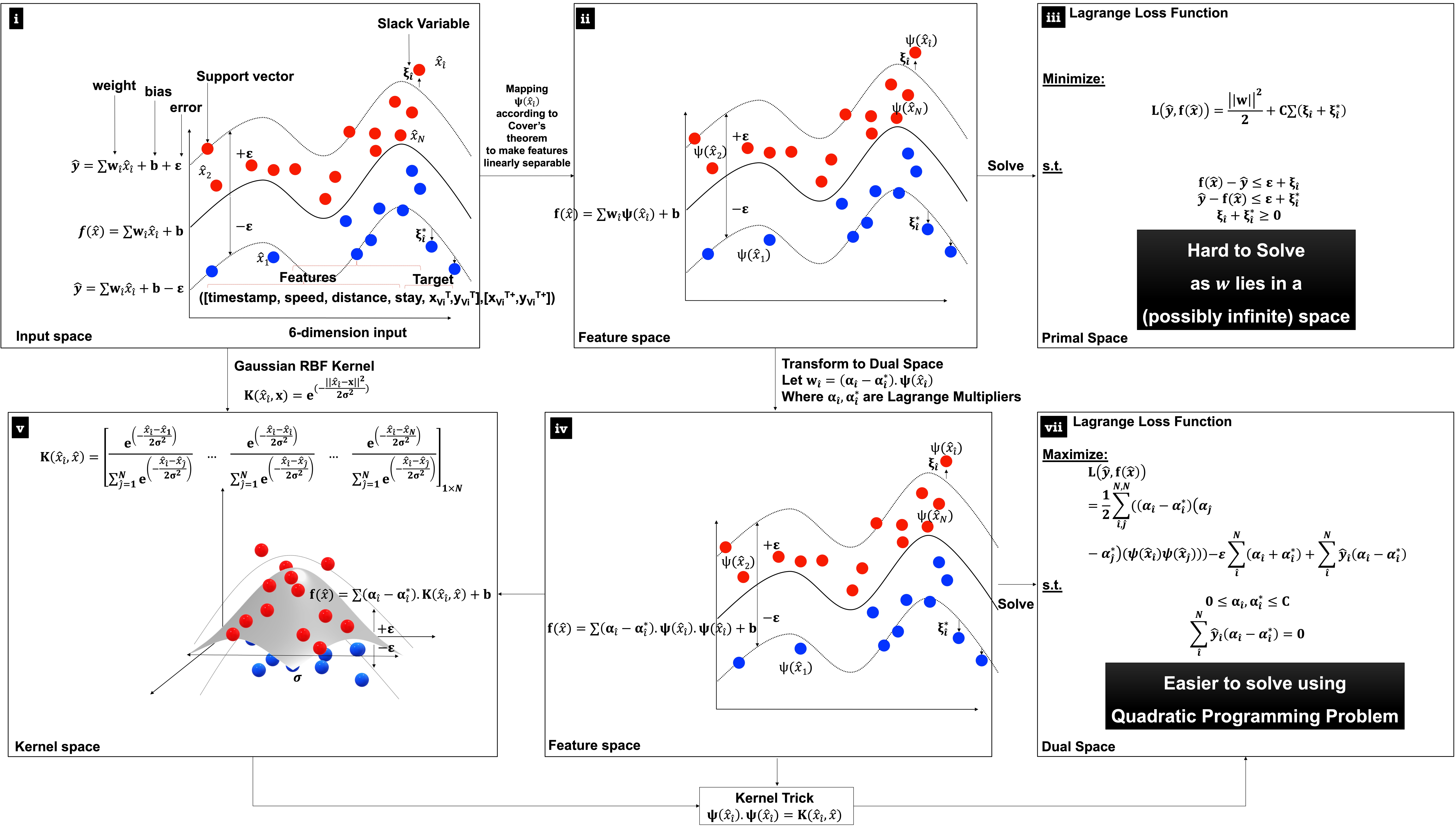}
    \caption{SVR Predictive Model Formulation}\label{fig:plan4}
\end{figure*}
\vspace{-5pt}
\begin{enumerate}[label=\roman*.]
\item In the input space, we identify the support vectors and the data points closest to the hyperplane to focus on the most critical data points when creating the prediction model:
\begin{equation}\label{eq:1}
f(\hat x)=\sum_{\hat i}^{N}(w_{\hat i}.{\hat x_{\hat i}})+b \pm \varepsilon,
\end{equation}
where $w_{\hat i}$ represents the weight, i.e., the normal to the optimal decision hyperplane of (\ref{eq:1}), and $b$ is the bias or the closest distance to the origin of the coordinate system. In this step, the slack variables $\xi_{\hat i}$ and $\xi_{\hat i}^*$ represent the allowable margin of error beyond the $\varepsilon$-tube.
\item To make the features of the input linearly separable, a mapping function $\psi(\hat x_{\hat i})$ is used instead of each $\hat x_{\hat i}$.
\item In such transformed feature space of $\psi(\hat x_{\hat i})$, the optimal hyperplane $f(\hat x)$ can be found by minimizing the following Lagrange Loss primal problem:
\begin{equation}\label{eq:2}
L(\hat y, f(\hat x))= \frac{{||w||}^2}{2}+C\sum{\xi_{\hat i}+\xi_{\hat i}^*},
\end{equation}
such that $C$ controls the generalization capabilities of the predictor, $\hat y - f(\hat x) \leq \varepsilon+\xi_{\hat i}$, $f(\hat x)-\hat y \leq \varepsilon+\xi_{\hat i}^*$, and $\xi_{\hat i}+\xi_{\hat i}^*\geq 0$ for $\hat i=\{1, ..., N\}$. As this problem (\ref{eq:2}) is hard to solve in this primal space of $w$, the solution is to transform the problem to a dual space of $\alpha_{\hat i}, \alpha^{*}_{\hat i}$ Lagrange multipliers by letting:
\begin{equation}\label{eq:3}
w_{\hat i}=(\alpha_{\hat i}-\alpha_{\hat i}^*).\psi(x_{\hat i}), \hat{i}=\{1, ..., N\}.
\end{equation}
\item In the dual feature space, the hyperplane is exhibited as:
\begin{equation}\label{eq:4}
f(\hat x)=\sum_{\hat i}^{N}(\alpha_{\hat i}-\alpha_{\hat i}^*).(\psi(\hat x_{\hat i}).\psi(\hat x_{\hat i})),
\end{equation}
where $(\psi(\hat x_{\hat i}).\psi(\hat x_{\hat i}))$ is the dot product between the two mapping functions. It turns out that the calculation of such a dot product can be avoided by using the kernel trick concept such that:
\begin{equation}\label{eq:5}
(\psi(\hat x_{\hat i}).\psi(\hat x_{\hat i}))=K(\hat x_{\hat i},\hat x), \hat i=\{1, ..., N\}.
\end{equation}
\item In the kernel space, from (\ref{eq:5}), the kernel trick can directly calculate the similarity between input features by transforming them into a higher-dimensional space. As the choice of the kernel function influences the flexibility and performance of the SVR model and as our dataset is a mix of linear and nonlinear features, we leverage the Gaussian Radial Basis Function (RBF) kernel to simplify capturing the nonlinear relationships for prediction. The formula for the RBF kernel in our SVR is $K(\hat x_{\hat i}, \hat x) = \exp \left( -\frac{||\hat x_{\hat i} - \hat x||^2}{2\sigma^2} \right)$, where $||\hat x_{\hat i} - \hat x||^2$ is the squared Euclidean distance and $\sigma$ is the width of the kernel's bell-shaped curve. A smaller $\sigma$ makes the kernel more localized, while a larger $\sigma$ makes it more spread out with potentially fewer support vectors; selecting an appropriate $\sigma$ requires cross-validation to find the optimal value for our specific dataset. As $\hat x$ has $N$ samples, we would calculate the kernel value for every $\hat x_{\hat i}$. This results in a $N \times N$ dimension kernel Gram matrix. From such a kernel matrix, we create a correlation matrix by subtracting the mean of each row and column from the matrix, ensuring that the kernel matrix has a zero mean. Next, we divide each element of the centered kernel matrix by the product of the square root of the corresponding diagonal elements. The correlation matrix provides insights into the relationships between data points in the higher-dimensional space as captured by the kernel function. Elements close to $1$ indicate high similarity, values close to $-1$ indicate anti-correlation, and values close to $0$ indicate low or no correlation.
\item Finally, in the dual space using the kernel trick, the Lagrange Loss function can be easily solved with Quadratic Programming to maximize:
\begin{equation}\label{eq:7}
\begin{split}
L(\hat y, f(\hat x))= \frac{1}{2}\sum_{\hat i, \hat j}^{N, N}((\alpha_{\hat i}-\alpha_{\hat i}^*).(\alpha_{\hat j}-\alpha_{\hat j}^*))K(\hat x_{\hat i},\hat x_{\hat j})\\-
\varepsilon\sum_{\hat i}^{N}(\alpha_{\hat i}+\alpha_{\hat i}^*)+\sum_{\hat i}^{N}\hat y_{\hat i}(\alpha_{\hat i}-\alpha_{\hat i}^*),
\end{split}
\end{equation}
such that $0\leq \alpha_{\hat i}, \alpha_{\hat i}^* \leq C$ and $\sum_{\hat i}^{N}\hat y_{\hat i}(\alpha_{\hat i}-\alpha_{\hat i}^*)=0$.
\end{enumerate}

Once this Lagrangian optimization is solved, that is, the optimal $\varepsilon$-tube hyperplane $f(\hat x)$ that best fits the training data $\hat x$ is found, we can use it to make predictions on new unseen data points.
\subsubsection{DNN DL Predictive Model}
The adopted DNN model is carefully designed to apprehend intricate spatiotemporal patterns inherent in our $\hat x$ dataset. The input layer of our model consists of six neurons to receive each $N$ 6-dimensional training and validation sample. We opt for a model of two hidden layers with 64 neurons and a Rectified Linear Unit (ReLU) activation function. The final layer has two neurons, each assigned to forecasting the x-coordinate and y-coordinate of an ambulance's next location. This architectural arrangement equips the DNN to learn and generalize the training data effectively. The model undergoes 1000 training iterations with a batch size of 32 out of the $N$ sample; 20\% of the training data serves as a validation subset. These settings balance model complexity and generalization, determined by empirical exploration. Once the model is finely tuned and trained, real-time data, including current vehicle location, is fed into the input layer. The model aims to precisely predict the future coordinates of the vehicle $(x_{V_i}^{T+}, y_{V_i}^{T+})$. This prediction is an invaluable tool for alleviating the observed delay $\Delta \tau$ and improving the overall efficiency of DT-HITS, as discussed in Section \ref{results}.

\section{Prediction Effect and DT Exhibition: Analysis and Discussion}\label{results}
In this section, our objective is to evaluate the performance of our SVR and DNN prediction models, providing a detailed account of their accuracy. We begin with an analysis of model evaluation metrics, shedding light on the precision and reliability of our models. Next, we showcase the models' proficiency in capturing underlying patterns in the testing geospatial data through accurately predicted values aligning with actual testing data. We then exhibit the mock DT implementation, showing the original and predicted locations in real time. Furthermore, we explore how our predictions offset the observed delay $\Delta \tau$, drawing comparisons between the observed improvement. We implemented our models and conducted our analysis and simulations in Python and MATLAB R2022b environments on a computer with a 2.8GHz Core i7 processor and 16GB memory.
\subsection{Predictive Models Validation Accuracy}
Measuring the accuracy and error of our two prediction models is crucial to assessing their effectiveness. In our analysis, we use the following three metrics for the assessment.
\begin{itemize}
\item Mean Absolute Error (MAE) calculates the average absolute differences between actual and predicted next location coordinates as:
\begin{equation}
MAE = \frac{1}{N}\sum_{\hat i=1}^{\hat N}(|(x_{V_i}^{T+})_{\hat i} - (\hat x_{V_i}^{T+})_{\hat i}| + |(y_{V_i}^{T+})_{\hat i} - (\hat y_{V_i}^{T+})_{\hat i}|)
\end{equation}

\item Mean Squared Error (MSE) squares the differences, placing more weight on larger errors.
\begin{equation}
MSE = \frac{1}{N}\sum_{\hat i=1}^{\hat N}(((x_{V_i}^{T+})_{\hat i} - (\hat x_{V_i}^{T+})_{\hat i})^2 + ((y_{V_i}^{T+})_{\hat i} - (\hat y_{V_i}^{T+})_{\hat i})^2)
\end{equation}

\item R-squared ($R^2$) measures how well our model's predictions explain the variability in the actual data. A value of $1$ indicates a perfect fit, while a value of $0$ indicates that the model does not explain any variability:
\begin{equation}
R^2 = 1 - \frac{\sum_{\hat i=1}^{\hat N}((x_{V_i}^{T+})_{\hat i} - (\hat x_{V_i}^{T+})_{\hat i})^2 + ((y_{V_i}^{T+})_{\hat i} - (\hat y_{V_i}^{T+})_{\hat i})^2}{\sum_{\hat i=1}^{\hat N}((x_{V_i}^{T+})_{\hat i} - \overline{{{\hat x}_{V_i}}^{T+}})^2 + ((y_{V_i}^{T+})_{\hat i} - \overline{{{\hat y}_{V_i}}^{T+}})^2}
\end{equation}
\end{itemize}

In the context of the evaluation metrics described, Table \ref{table:result1} offers a comprehensive analysis of the accuracy performance for our SVR and DNN models in two distinct computational environments. MATLAB and Python. Within the MATLAB environment, the SVR RBF model exhibits a moderately high MAE of $83.424$ and MSE of $15527.804$, suggesting a moderate level of predictive accuracy. However, $R^2$ stands impressively high at $0.99911$, indicating a robust correlation between predicted and actual values. In contrast, the DNN model in MATLAB outperforms the SVR RBF, achieving a significantly lower MAE of $9.179$ but with a higher MSE of $17261.584$. Despite a slightly less perfect $R^2$ value of $0.99901$, this underscores the superior predictive accuracy of the SVR RBF model and its ability to capture intricate patterns within the validation dataset. SVR RBF and DNN models demonstrate remarkable accuracy when transitioning to the Python environment. The SVR RBF achieves an MAE of $0.0712$ and an MSE of $0.0265$, with a highly recommended value of $R^2$ of $0.97345$, indicating a precise alignment with the validation data. The DNN model in Python excels further, achieving a value of $R^2$ of $0.99995$, accompanied by negligible MAE ($0.0105$) and moderate MSE ($0.0002$), highlighting its exceptional predictive accuracy and its ability to capture the underlying patterns within the validation data set accurately. Consistently across both environments, the DNN model outperforms the SVM RBF, with Python implementations yielding superior results. The consistently high values $R^2$ across all models affirm their reliability and suitability for predictive tasks in MATLAB and Python environments.

\begin{table}[!htbp]
\centering \caption{Accuracy Performance Metrics on Validation Dataset}\label{table:result1}
\begin{tabular}{|c|c|c|c|c|c|}\hline
\textbf{Model Type} & \textbf{Environment} & \textbf{MAE} & \textbf{MSE} & \textbf{R$^2$} \\\hline
SVM RBF & MATLAB &83.424&15527.804&0.99911 \\\hline
DNN & MATLAB &9.179&17261.584&0.99901 \\\hline
SVM RBF &Python&0.0712&0.0265&0.97345\\\hline
DNN &Python&0.0105&0.0002&0.99995\\\hline
\end{tabular}
\end{table}
\vspace{-5pt}
\subsection{Prediction Testing Scenarios Results}
\begin{figure*}[!htbp]
\centering
    \begin{subfigure}[b]{0.2\linewidth}
        \centering
        \includegraphics[width=\textwidth]{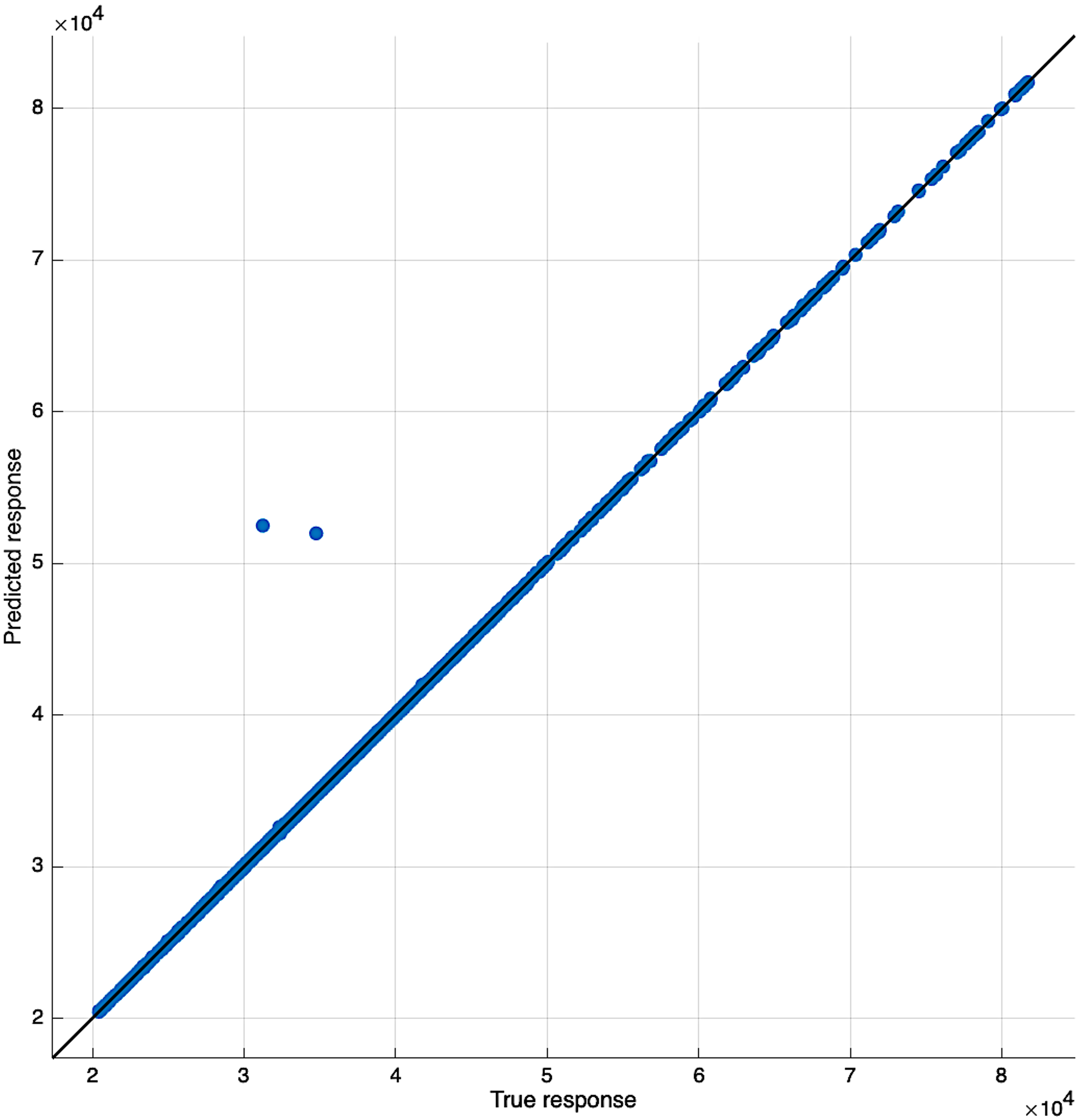}
        \caption{DNN: Latitude}\label{fig:v1}
    \end{subfigure}%
    \begin{subfigure}[b]{0.2\linewidth}
        \centering
        \includegraphics[width=\textwidth]{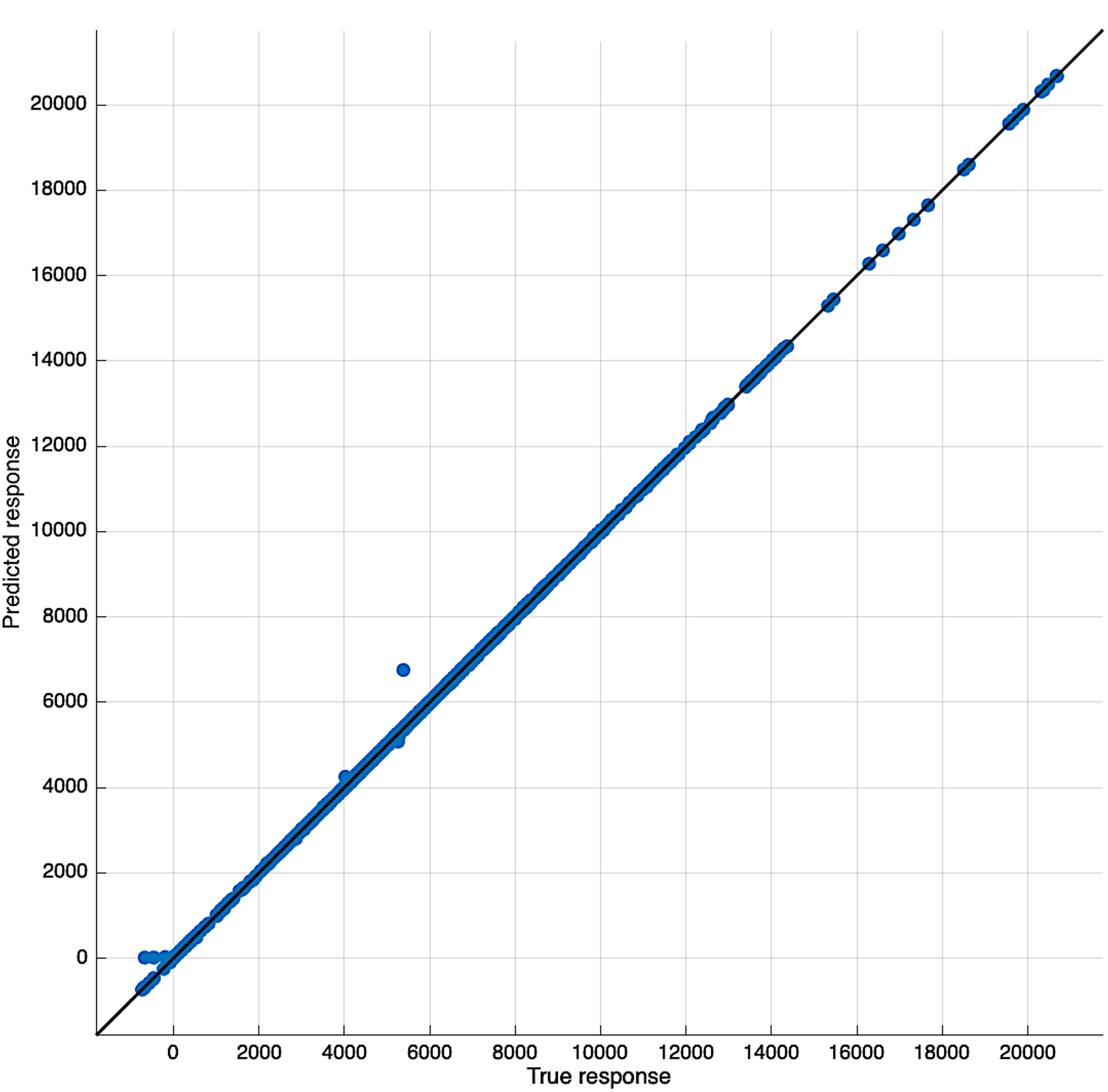}
        \caption{DNN: Longitude}\label{fig:v2}
    \end{subfigure}
    \begin{subfigure}[b]{0.2\linewidth}
        \centering
        \includegraphics[width=\textwidth]{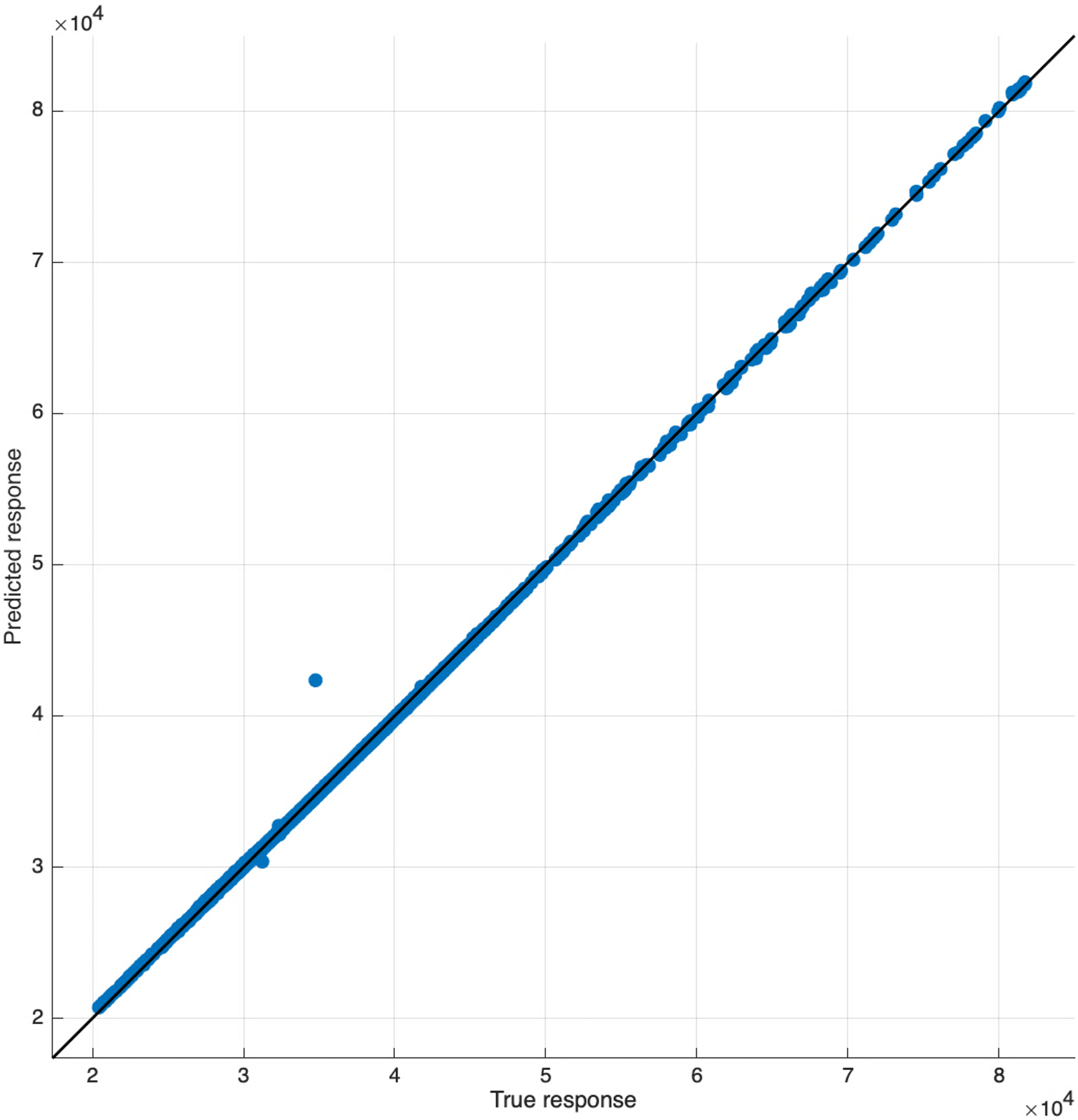}
        \caption{SVR: Latitude}\label{fig:v3}
    \end{subfigure}%
    \begin{subfigure}[b]{0.2\linewidth}
        \centering
        \includegraphics[width=\textwidth]{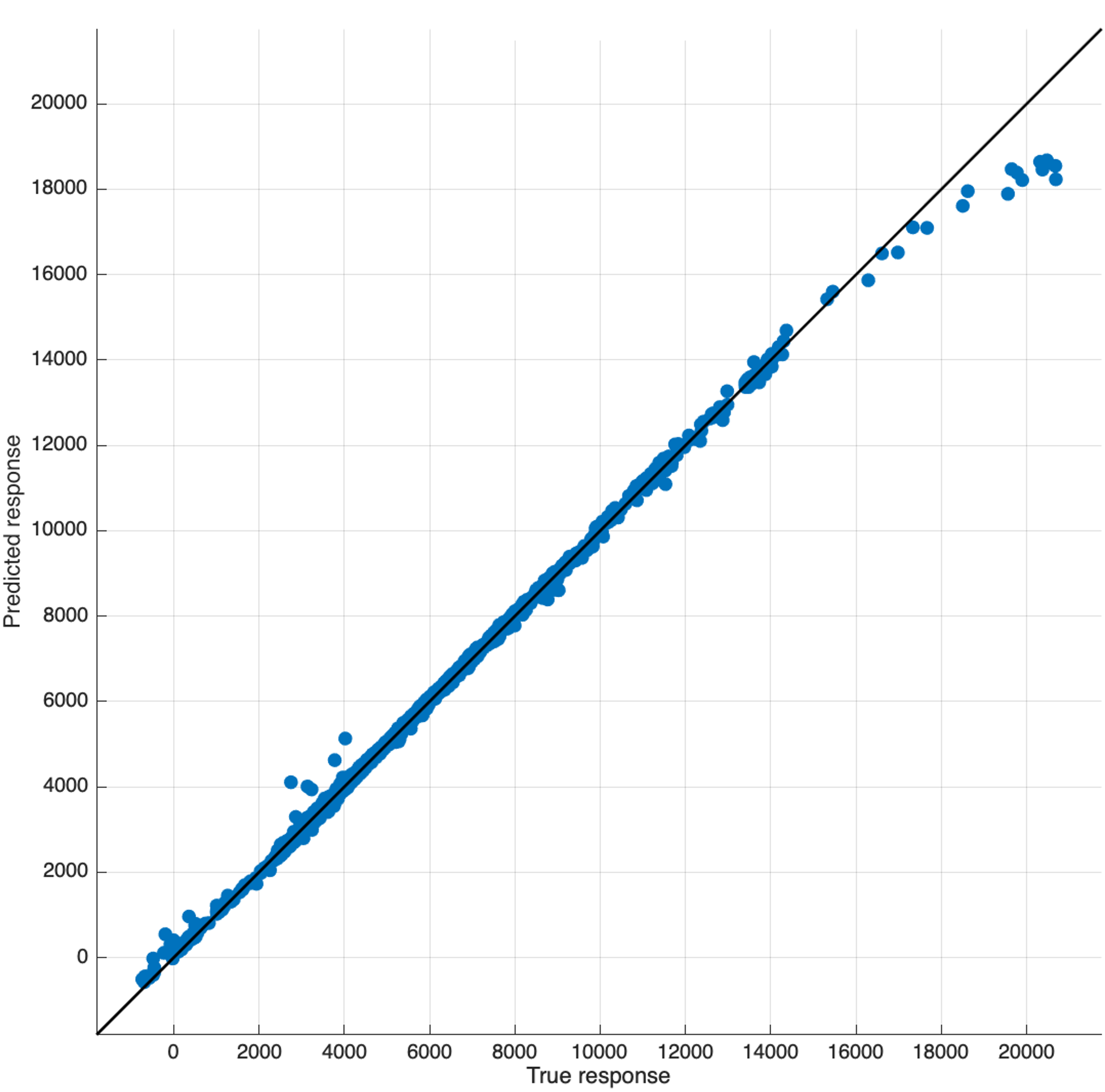}
        \caption{SVR: Longitude}\label{fig:v4}
    \end{subfigure}
    \caption{DNN and SVR Models Scenario 1's True vs Predicted Response in MATLAB}
    \label{fig:testmatlab1}
\end{figure*}

\begin{figure*}[!htbp]
\centering
    \begin{subfigure}[b]{0.2\linewidth}
        \centering
        \includegraphics[width=\textwidth]{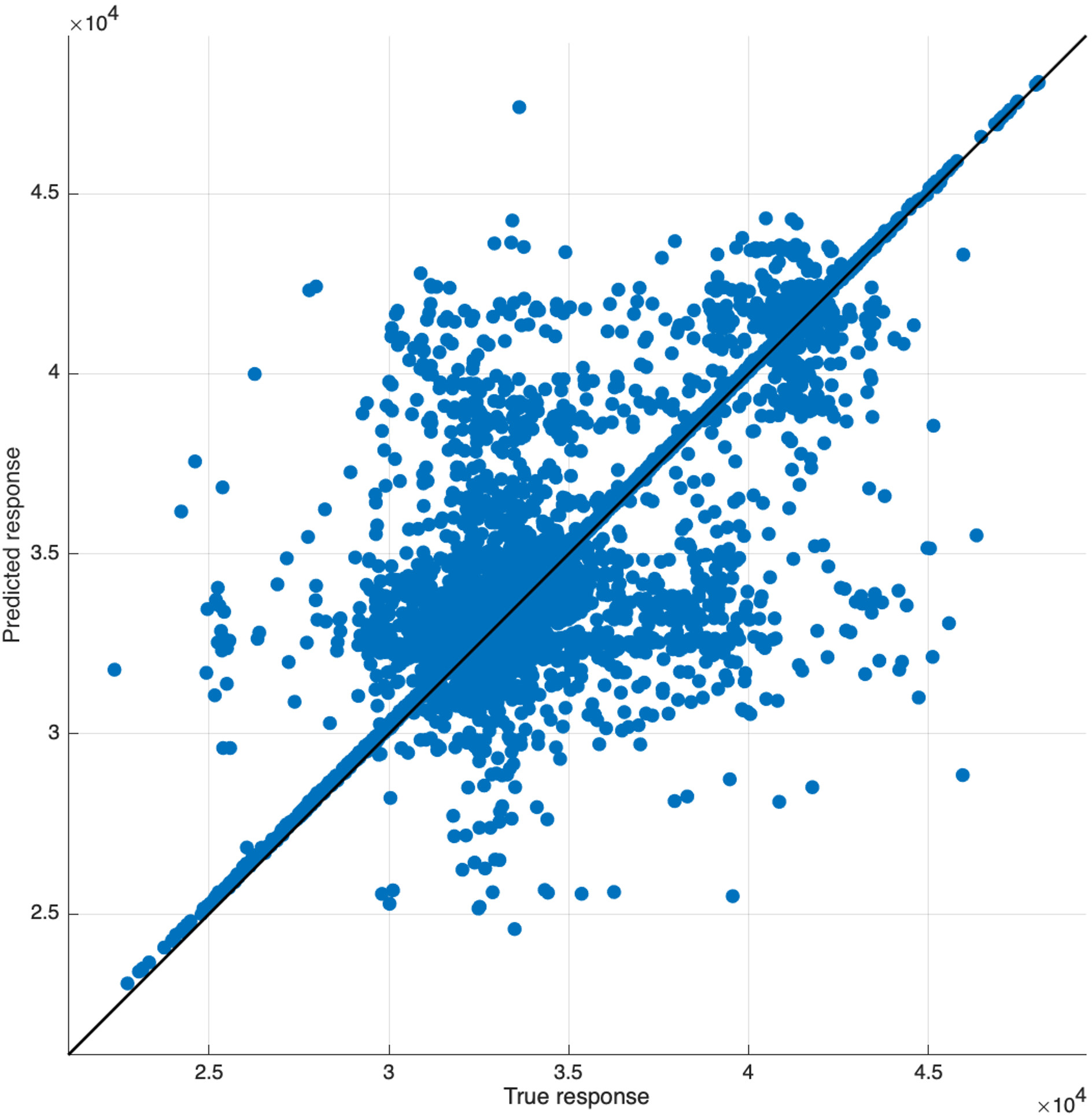}
        \caption{DNN: Latitude}\label{fig:v11}
    \end{subfigure}%
    \begin{subfigure}[b]{0.2\linewidth}
        \centering
        \includegraphics[width=\textwidth]{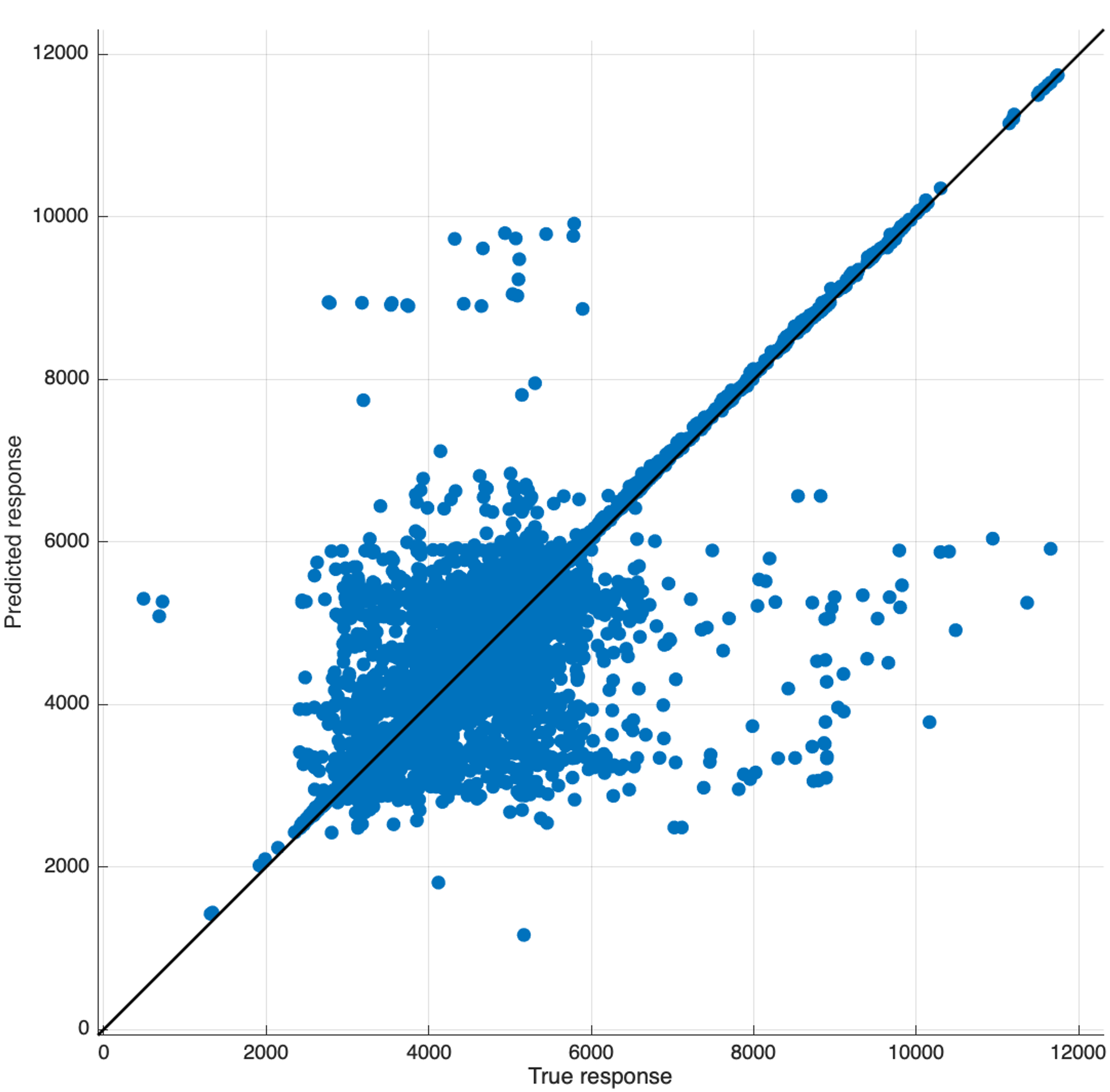}
        \caption{DNN: Longitude}\label{fig:v22}
    \end{subfigure}
    \begin{subfigure}[b]{0.2\linewidth}
        \centering
        \includegraphics[width=\textwidth]{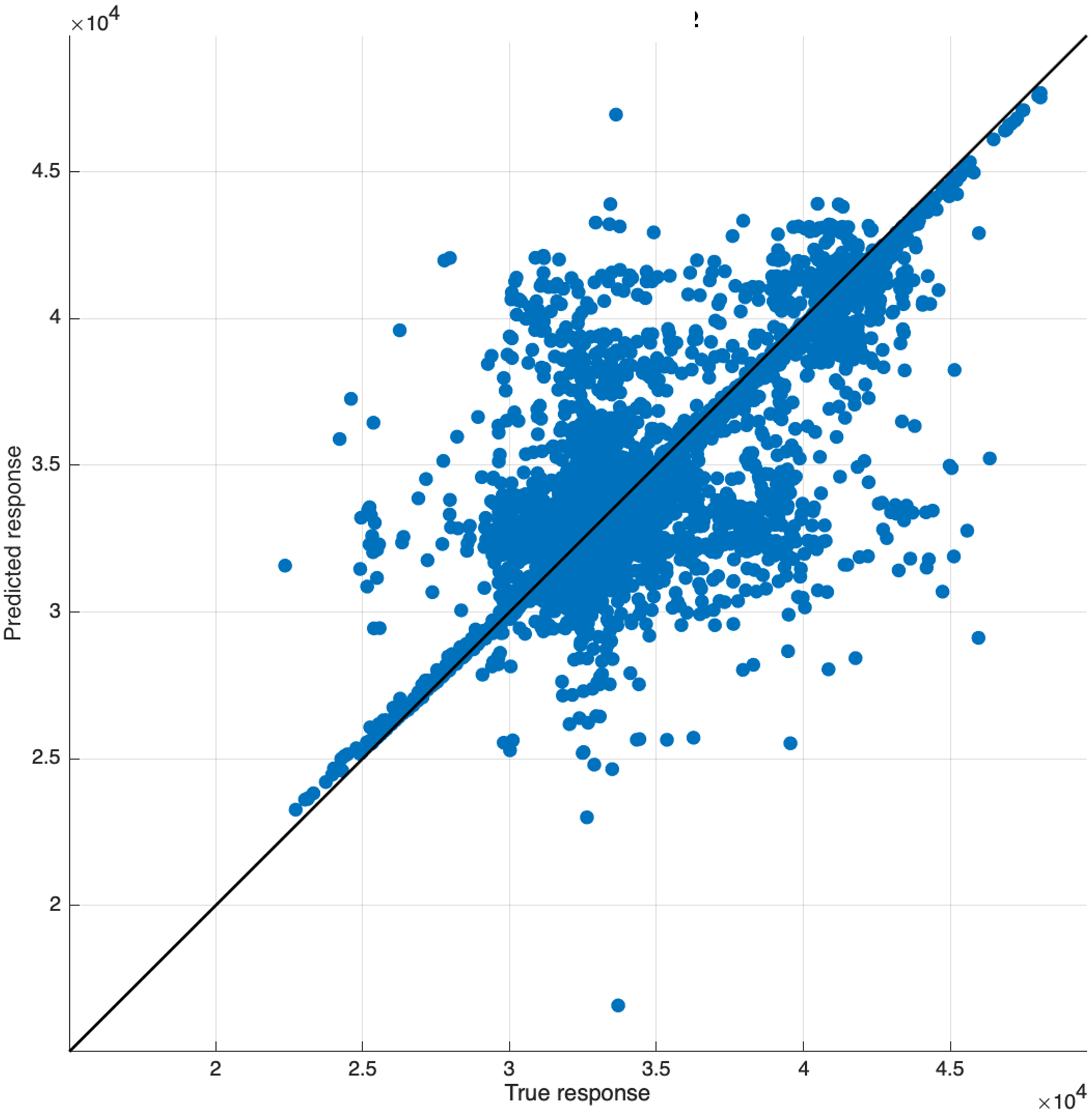}
        \caption{SVR: Latitude}\label{fig:v33}
    \end{subfigure}%
    \begin{subfigure}[b]{0.2\linewidth}
        \centering
        \includegraphics[width=\textwidth]{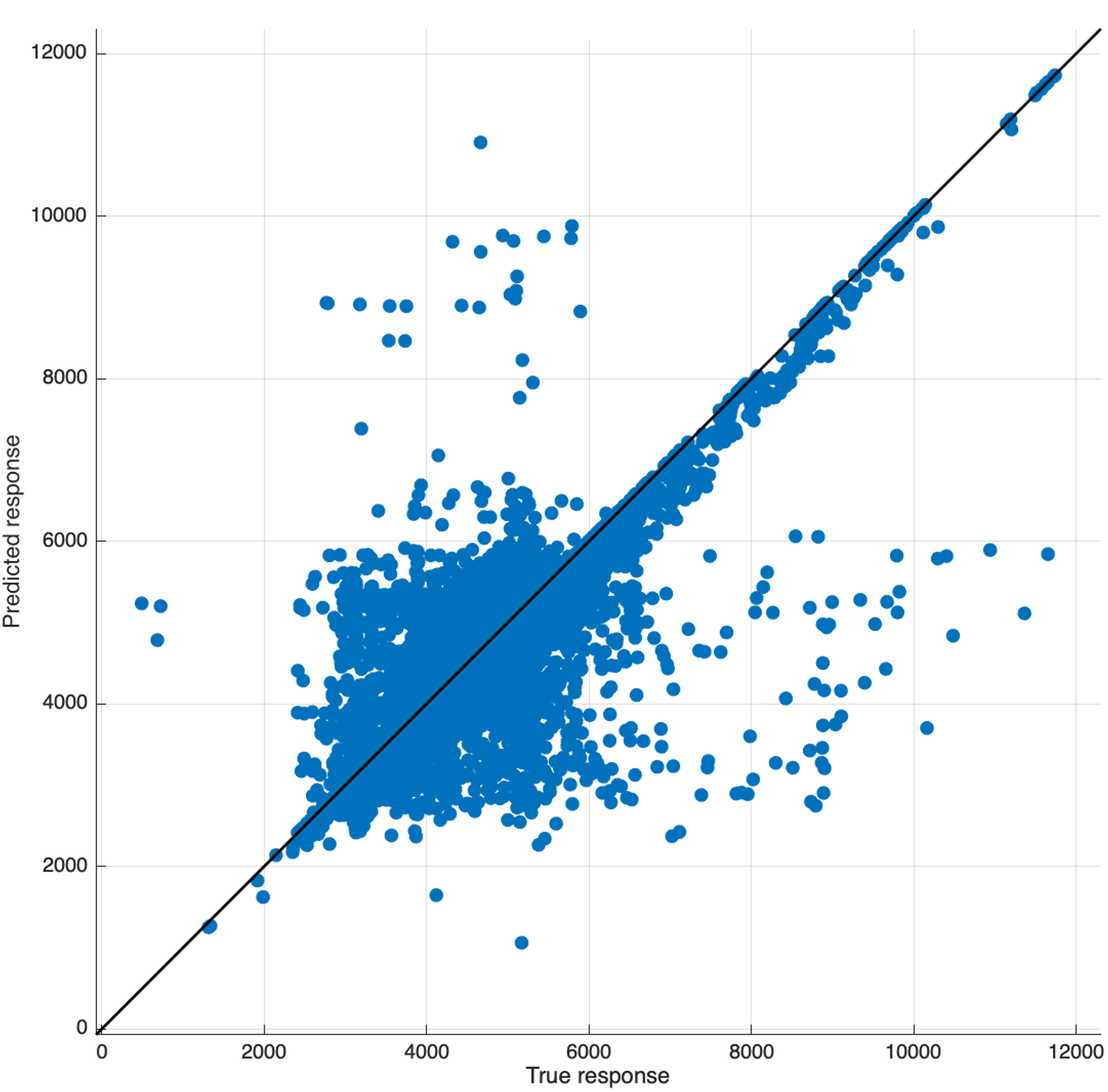}
        \caption{SVR: Longitude}\label{fig:v44}
    \end{subfigure}
    \caption{DNN and SVR Models Scenario 2's True vs Predicted Response in MATLAB}
    \label{fig:testmatlab2}
\end{figure*}

In evaluating the generalization of the models to unseen data, two datasets denoted $\hat x$ are used with 40,128 samples (Scenario 1) and 19,252 samples (Scenario 2), each with six-dimensional input features. Testing is carried out in MATLAB and Python environments to ensure cross-platform consistency. MATLAB-visualized results, Fig. \ref{fig:testmatlab1}, reveal that for the larger dataset of Scenario 1, the predicted latitude and longitude responses are closely aligned with the true responses, though there are occasional outliers. However, the test results for the smaller dataset in Scenario 2, Fig. \ref{fig:testmatlab2}, indicate a higher error. This discrepancy suggests potential challenges in the models' ability to generalize effectively to datasets with fewer samples, offering valuable insights into their performance across varying data sizes.

\begin{figure}[!htbp]
\centering
    \begin{subfigure}[b]{.8\linewidth}
        \centering
        \includegraphics[width=\textwidth]{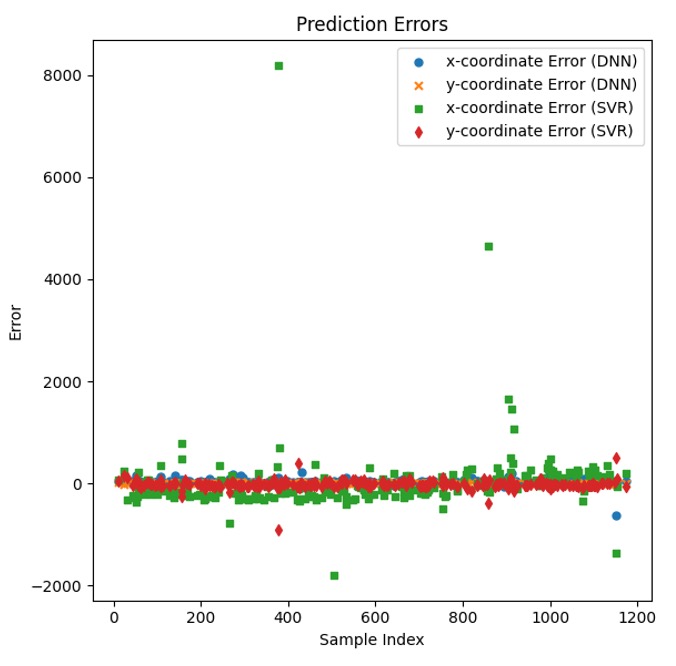}
        \caption{Scenario 1}\label{fig:pythonsc1}
    \end{subfigure}%
    
    \begin{subfigure}[b]{.8\linewidth}
        \centering
        \includegraphics[width=\textwidth]{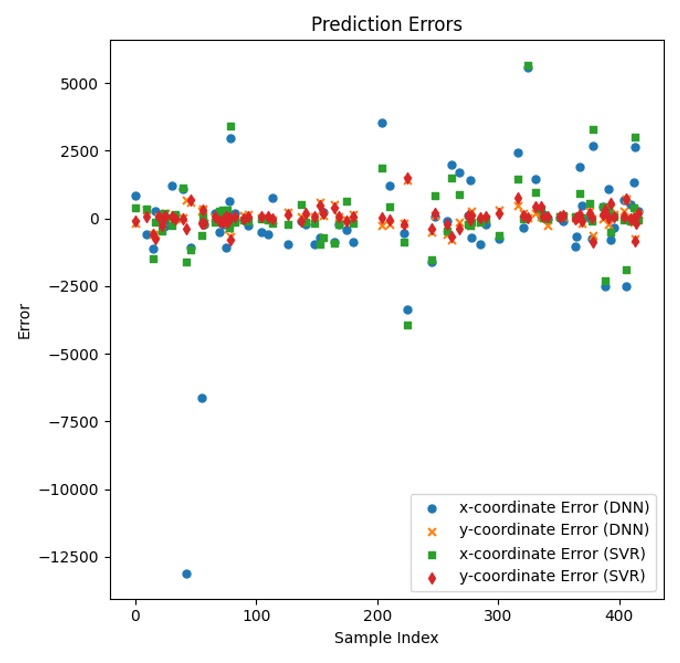}
        \caption{Scenario 2}\label{fig:pythonsc2}
    \end{subfigure}
\caption{SVR and DNN Models' True vs. Predicted Next Locations and Reported Errors in Python}\label{fig:pythonsc12}
\end{figure}

The Python results exhibit a more favorable performance in predicting location than the actual location of the emergency vehicle for both test datasets, as shown in Fig. \ref{fig:pythonsc12}. These visualizations suggest a higher accuracy in the Python environment, showcasing a closer alignment between the predicted and true responses for the ambulance's next location. The implication is that, in contrast to the MATLAB results discussed earlier, the Python implementation of the models shows superior performance in predicting the ambulance's location, highlighting the programming environment's importance in influencing the models' effectiveness.

\subsection{Showcasing The Prediction in The DT}
\begin{figure}[!htbp]
    \centering
    \includegraphics[width=\linewidth]{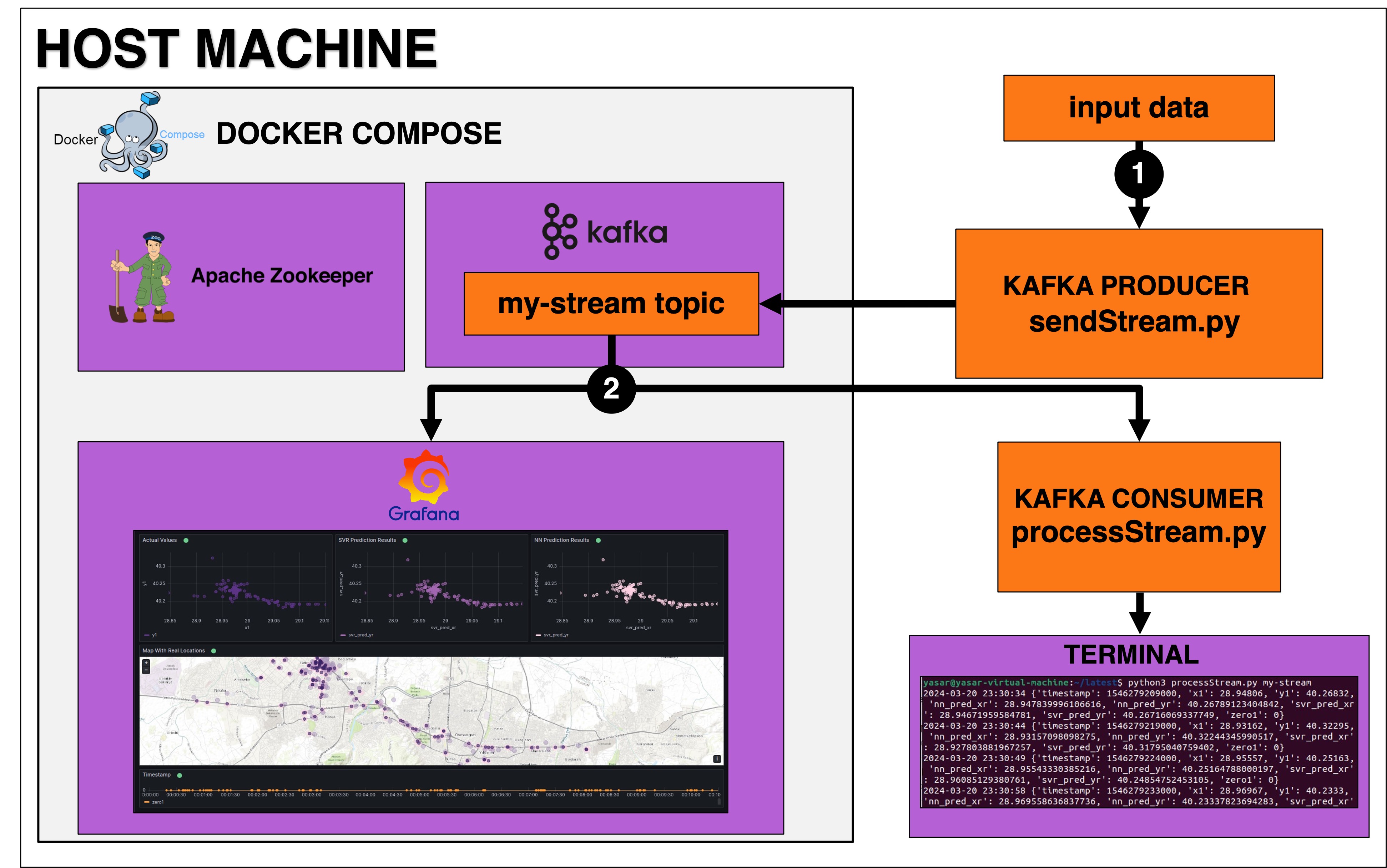}
    \caption{Mock DT for Data Pipeline Architecture}\label{fig:mockdt}
\end{figure}

A mock DT environment has been developed on the host machine\footnote{We build the visualization environment on a virtual machine that runs the Ubuntu 22.04 operating system and has 4GB memory.} to emulate the original and predicted ambulance locations. Using the open-source Docker platform for containerization, the DT application and its associated components and dependencies are encapsulated within virtual containers; a Docker container is constructed using a Docker package/image \cite{Docker}. The orchestration of these containers is adeptly managed by Docker-compose, which is crucial for the cohesive operation of the DT system. The resultant integrated DT system is contained within the Docker-compose framework, as illustrated in Fig. \ref{fig:mockdt}. Three main images are used. For data streaming, Apache Kafka, renowned for its scalability and fault tolerance capabilities, is set by the "confluentinc/cp-kafka:latest" image \cite{kafka}. Apache Zookeeper is used through the "confluentinc/cp-zookeeper:latest" image to coordinate and synchronize the Kafka streaming nodes (brokers) to ensure consistent management of system coordination \cite{Zookeeper}. Lastly, the Grafana visualization, represented by the "grafana/grafana:latest" image, builds a connection agent to integrate Kafka with the visualization, i.e., to send Kafka metrics to the Grafana Cloud instance.

\begin{figure*}[!htbp]
    \centering
    \includegraphics[width=\linewidth]{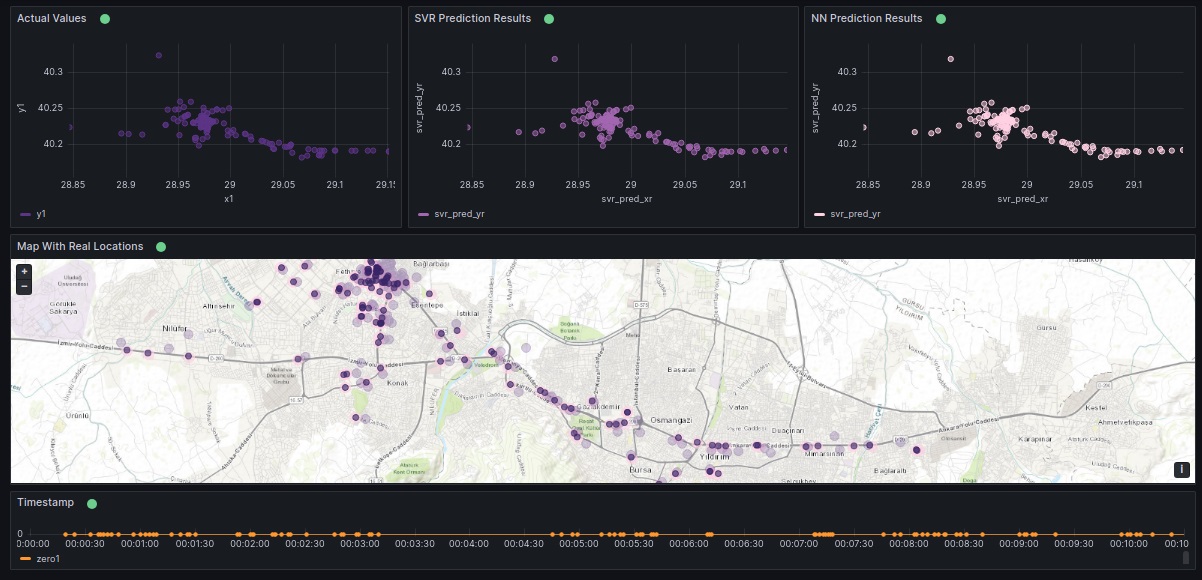}
    \caption{Developed HITS DT: Original vs. Predicted Locations (for a recorded visualization excerpt see \cite{celik2024mock})}\label{fig:mockdtprediction}
\end{figure*}

The data pipeline in our DT starts with reading the input data. We used a CSV file with timestamp, latitude, and longitude coordinates of the current location $(x_{V_i}^T, y_{V_i}^T)$, the predicted coordinates of the next corresponding location $(x_{V_i}^{T+}, y_{V_i}^{T+})$ using SVR and DNN for an ambulance of the 221 vehicles in the original data set. A Python script named "sendStream.py" parses the CSV file. This script initializes the Kafka Producer, which is constructed using the "confluent-Kafka" library, a toolset that facilitates the employment of Kafka functionalities within the Python environment. The Kafka Producer sends the read data to the Kafka topic, which serves as a fundamental organizational entity within the Kafka cluster, enabling Producers to dispatch data and Kafka Consumers to retrieve data from this communication medium. In this setup, the Kafka Producer processes each row of the CSV file, translating it into a JSON-formatted message before sending it to the'my-stream' topic in Kafka as delineated in Algorithm \ref{algo:1}. The script calculates the temporal interval between successive timestamps to simulate a live data stream. Consequently, it introduces a delay in transmitting messages, ensuring that the temporal fidelity of the data is maintained.
Conversely, the consumer-side Python script, "processStream.py," is responsible for instantiating the 'my-stream' topic and the Kafka Consumer. This consumer is configured to subscribe to the 'my-stream' topic and outputs the incoming messages to the terminal. Integration with Grafana is achieved through the use of an Apache Kafka plugin \cite{grafana-kafka-plugin}, which registers the Kafka Broker as a data source within Grafana, thereby granting it access to the data streamed to the "my-stream" topic, as explained in Algorithm \ref{algo:2}. Our Grafana, Fig. \ref{fig:mockdtprediction} is configured to create five different visualization panels. These panels display longitude-latitude charts representing the testing data locations, the predictions from SVR and DNN, a geomap that plots the data, and a temporal visualization of the timestamps.

\begin{algorithm}\footnotesize
\caption{Send data to Kafka topic \cite{celik2024mock}}\label{algo:1}
\begin{algorithmic}
\Function{send\_to\_kafka\_topic}{input\_data}
    \State Initialize KafkaProducer
    \For{each data\_point in input\_data}
        \State KafkaProducer.send('my-stream-topic', value=data\_point)
    \EndFor
    \State Close KafkaProducer
\EndFunction
\end{algorithmic}
\end{algorithm}

\begin{algorithm}\footnotesize
\caption{Consume data from Kafka topic \cite{celik2024mock}}\label{algo:2}
\begin{algorithmic}
\While{True}
    \Function{consume\_from\_kafka\_topic}{}
        \State Initialize KafkaConsumer for 'my-stream-topic'
        \For{each message in KafkaConsumer}
            \State data\_point = message.value
            \State Process data\_point
            \State Store or forward processed data for visualization
        \EndFor
    \EndFunction
\EndWhile
\end{algorithmic}
\end{algorithm}

\subsection{Witnessed Delay Offset}
To assess the impact of prediction on mitigating the observed delay ($\Delta \tau$) between the digital and physical domains, our simulation-based analysis initially examines the communication delay observed from the perspective of the Road Side Unit (RSU) in the physical network. With an RSU coverage of 1 km that accommodates around $n=40$ vehicles, the RSU uses V2I communication through a cellular network with a data transfer rate of 100 Mbps. Furthermore, V2V communication employs WiFi at 6Mbps, featuring a control channel duration of 46 msec. Specifically, for a safety application that transmits 310 bytes of data, a vehicle beacon rate of 100 msec, and an application processing time of 2.23 msec, Table \ref{table:delay} illustrates a noticeable correlation between the increase in the number of vehicles $n$ and the increase in delay in the physical HITS realm. Upon deploying virtual twin prediction models for this HITS at the RSU's edge server, improvements in the observed temporal gap ($\Delta \tau$) between the DT and the physical system are evident, as depicted in Table \ref{table:delay}, including the average testing prediction delays of 0.0883 sec for the DNN model and 0.0037 sec for the SVR model. The results in Table \ref{table:delay} show the efficacy of using predictive models to bridge the temporal gap in HITS, especially as the number of vehicles increases. The DNN and SVR models contribute to substantial $\Delta \tau$ reductions, reflecting improved synchronization between the digital and physical realms.

\bgroup
\begin{table}[!htbp]
\centering
\caption{Witnessed Communication Delay $\Delta \tau$ (sec): when no DT is used and when DT with prediction is used}\label{table:delay}
\renewcommand{\arraystretch}{1}
\footnotesize
\begin{tabular}{|l|l|l|l|}\hline
$n$&\textbf{No DT}&\textbf{DT and Prediction}&\textbf{Improvement(\%)}\\ \hline
2 & 1.657793333&0.196686667&88.13\\\hline
5 & 4.144483333 &0.347716667&91.61\\\hline
10 & 8.288966667 &0.599433333&92.76\\\hline
15 & 12.43345 &0.85115&93.15\\\hline
20 & 16.57793333 &1.102866667&93.34\\\hline
25 & 20.72241667 &1.354583333&93.46\\\hline
30 & 24.8669 &1.6063&93.54\\\hline
35 & 29.01138333 &1.858016667&93.59\\\hline
40 & 33.15586667 &2.109733333&93.63\\\hline
\end{tabular}
\end{table}
\egroup

\section{Conclusion and Key Extensions}\label{conc}
Despite the prevalent claim of real-time representation, the digital counterpart of HITS must always catch up with its physical world due to synchronization delays $\Delta \tau$. Our approach leveraged AI models to forecast the next-ambulance locations in the virtual world and align the virtual positions with their physical twin locations. We built a mock DT environment using Docker and Kafka to support a real-time data pipeline of actual and predicted locations; our DT enabled accurate visualization through Grafana, enhancing synchronization in HITS operations. In this DT, both the SVR and DNN models showcased high prediction accuracy in various testing scenarios and visually underscored the effectiveness of our methodology. In particular, DNN was the superior model, outperforming SVR in multiple instances. Significantly, both models played a transformative role by substantially reducing observed delays from 1.65 sec in the case of n=2 vehicles to only 0.196 sec and from 33 sec in the case of n=40 vehicles to only 2.1 sec. These results highlighted the efficacy of our proposed approach and emphasized the pivotal role of advanced AI predictive models in addressing such critical challenges. Future work could focus on ensemble modeling, combining SVR and DNN for improved predictions, and exploring hybrid solutions like edge-cloud DT integration for enhanced real-time processing efficiency.

\section*{Acknowledgment}
This work was supported by the USDOT UTC CARMEN + Project and the Turkish Scientific and Technological Research Council (TUBITAK) 1515 Frontier R$\&$D Laboratories Support Program for BTS Advanced AI Hub: BTS Autonomous Networks and Data Innovation Lab Project 5239903.

\bibliographystyle{unsrt}
\bibliography{bconf}
\end{document}